\documentclass[english]{IEEEtran}

\usepackage{cite}
\usepackage{amsmath,amssymb,amsfonts}
\usepackage{algorithmic}
\usepackage{graphicx}
\usepackage{textcomp}
\usepackage{xcolor}
\usepackage{subcaption}
\usepackage{xargs}
\usepackage{float}

\usepackage{hyperref}
\hypersetup{
    colorlinks,
    linkcolor={black},
    citecolor={blue!50!black},
    urlcolor={blue!80!black}
}
\usepackage[capitalise]{cleveref} 

\usepackage[ruled,vlined,linesnumbered]{algorithm2e}
\def\BibTeX{{\rm B\kern-.05em{\sc i\kern-.025em b}\kern-.08em
    T\kern-.1667em\lower.7ex\hbox{E}\kern-.125emX}}

\begin{document}

\title{Differentiable Nonparametric Belief Propagation}

\author{Anthony Opipari$^{1}$, Chao Chen$^{1}$, Shoutian Wang$^{1}$, Jana Pavlasek$^{1}$, Karthik Desingh$^{2}$, Odest Chadwicke Jenkins$^{1}$ 
\thanks{$^{1}$Robotics Institute, University of Michigan, Ann Arbor
        {\tt\small \{topipari,joecc,shoutian,pavlasek,ocj\}@umich.edu}}
\thanks{$^{2}$Department of Computer Science \& Engineering, University of Washington {\tt\small kdesingh@cs.washington.edu}}
}

\maketitle

\newcommand{\msg}[3]{m_{#1\rightarrow #2}^{#3}(X_{#2})}
\newcommand{\bel}[2]{bel_{#1}^{#2}(X_{#1})}
\newcommand{\prt}[3]{\mu_{#1#2}^{(#3)}}
\newcommand{\wgt}[3]{w_{#1#2}^{(#3)}}
\newcommand{\set}[1]{\{#1\}}

\newcommand{\unary}[3]{%
\ifthenelse{\equal{#2}{}}{\phi_{#1}(X_{#1},Y_{#1})}{\ifthenelse{\equal{#3}{}}{\phi_{#1}(X_{#1}=#2,Y_{#1})}{\phi_{#1}(X_{#1}=#2,Y_{#1}=#3)}}%
}
\newcommand{\pairwise}[4]{%
\ifthenelse{\equal{#3}{}}
{\psi_{#1,#2}(X_{#1},X_{#2})}
{\ifthenelse{\equal{#4}{}}
  {\psi_{#1,#2}(X_{#1},X_{#2}=#3)}
  {\psi_{#1,#2}(X_{#1}=#3,X_{#2}=#4)}}%
}

\newcommand{\temporal}[1]{\tau_{#1}(\epsilon)}

\global\csname @topnum\endcsname=1

\begin{abstract}
We present a differentiable approach to learn the probabilistic factors used for inference by a nonparametric belief propagation algorithm. Existing nonparametric belief propagation methods rely on domain-specific features encoded in the probabilistic factors of a graphical model. In this work, we replace each crafted factor with a differentiable neural network enabling the factors to be learned using an efficient optimization routine from labeled data. By combining differentiable neural networks with an efficient belief propagation algorithm, our method learns to maintain a set of marginal posterior samples using end-to-end training.
We evaluate our differentiable nonparametric belief propagation (DNBP) method on a set of articulated pose tracking tasks and compare performance with a recurrent neural network. Results from this comparison demonstrate the effectiveness of using learned factors for tracking and suggest the practical advantage over hand-crafted approaches. The project webpage is available at: \href{https://progress.eecs.umich.edu/projects/dnbp/}{progress.eecs.umich.edu/projects/dnbp}.
\end{abstract}

\section{Introduction}

A significant challenge for robotic applications is the ability to estimate the pose of articulated objects in high noise environments. Nonparametric belief propagation (NBP) algorithms~\cite{nbp:SudderthIFW03,isard2003pampas} have proven effective for inference in visual perception tasks such as human pose tracking~\cite{looselimbed:SigalBRBI04} and articulated object tracking in robotic perception~\cite{bpposeest:DesinghLOJ19, pavlasek2020parts}. Moreover, these algorithms are able to account for uncertainty in their estimates when environmental noise is high. Their adaptability to new applications, however, is limited by the need to define hand-crafted functions that describe the distinct statistical relationships in a particular dataset. Current methods that utilize NBP rely on extensive domain knowledge to parameterize these relationships~\cite{nbp:SudderthIFW03, sudderth2004visual, bpposeest:DesinghLOJ19, pavlasek2020parts, isard2003pampas, ihler2009particle}. Reducing the domain knowledge required by NBP methods would enable their use in a broader range of applications. 

As a form of probabilistic graphical model inference, NBP algorithms leverage domain knowledge encoded in graph-based representations, such as the Markov random field (MRF). Their capacity to perform inference using arbitrary graphs sets them apart from other algorithms such as the recursive Bayes filter~\cite{probrob:thrun} (e.g. particle filter~\cite{partfiltreview}) and has been shown to be important in computational perception because it allows for modeling of non-causal relationships~\cite{nbp:SudderthIFW03}. Data-driven approaches are an alternative for computational perception~\cite{xiang2018posecnn, dope:tremblay2018corl}. These methods generally avoid the need for extensive domain knowledge by learning from large amounts of labelled data. Data-driven approaches, however, are prone to noisy estimates and have limited capacity to represent uncertainty inherent in their estimates. In robotic applications, both of these limitations negatively impact the ability for a robot to operate effectively in unstructured environments.

\begin{figure}[t!]
\centering
\includegraphics[width=\columnwidth,keepaspectratio]{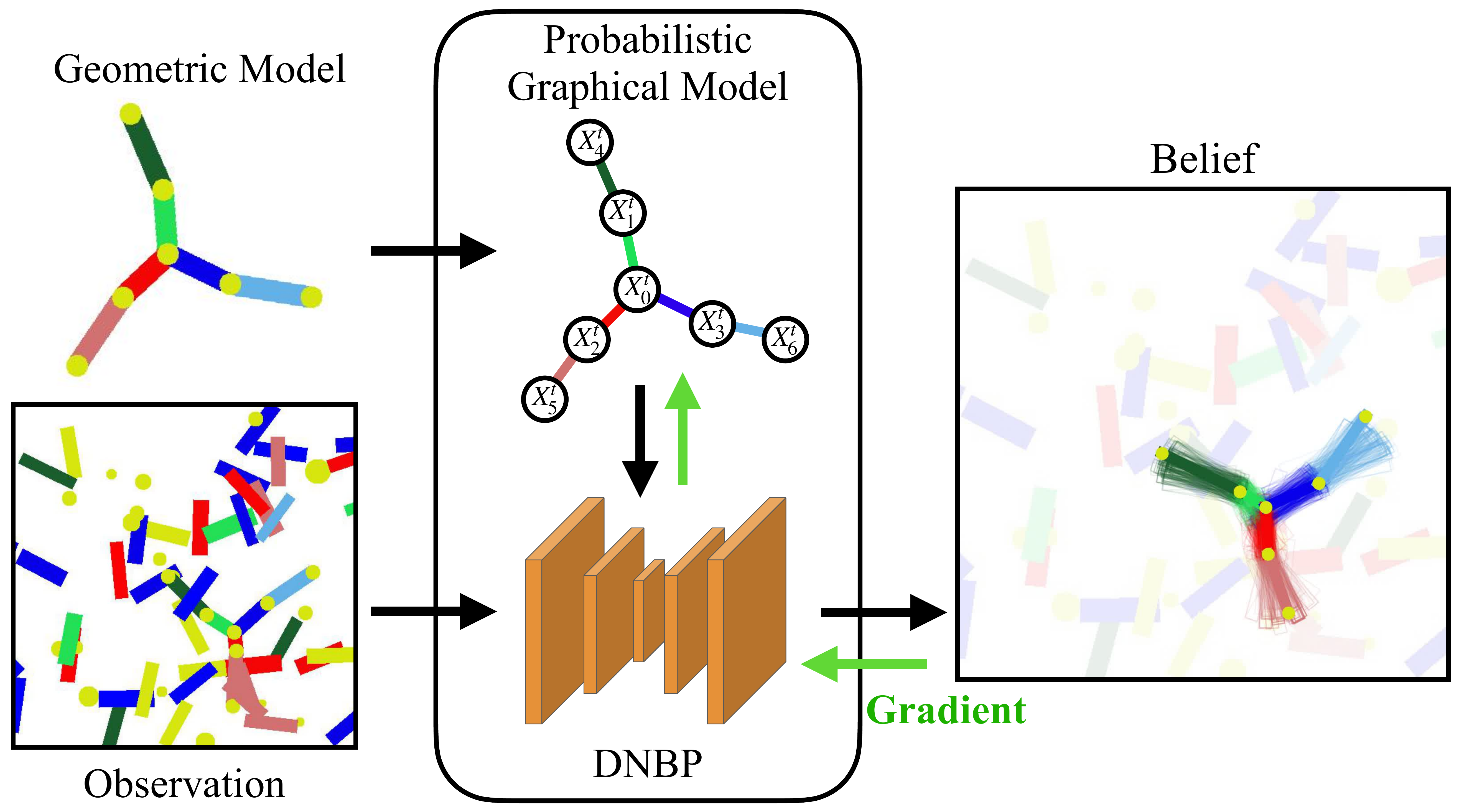}
\caption{Differentiable nonparametric belief propagation combines domain knowledge in the form of graphical models with differentiable neural networks. Probabilistic relationships encoded in the graphical model are learned end-to-end using backpropagation.}
\label{fig:teaser}
\end{figure}

In this paper, we present a differentiable nonparametric belief propagation (DNBP) method, a hybrid approach which leverages neural networks to parameterize the NBP algorithm. DNBP overcomes the need for hand-crafted functions by using data-driven techniques while retaining the strengths of probabilistic inference. Through differentiable inference, DNBP leverages the explainability and robustness of probabilistic inference techniques and capitalizes on the efficiency and generalizability of data-driven approaches. Specifically, we develop a differentiable version of the efficient pull message passing nonparametric belief propagation (PMPNBP) algorithm presented by Desingh et al.~\cite{bpposeest:DesinghLOJ19}. We are inspired by the differentiable particle filter proposed by Jonschkowski et al.~\cite{dpf:JonschkowskiRB18}. Similar to this approach, DNBP performs end-to-end learning of each probabilistic factor required for graphical model inference. 
DNBP has the potential to generalize broadly to high-dimensional articulated tracking tasks in challenging cluttered environments.

The effectiveness of DNBP is demonstrated on two simulated articulated tracking tasks in challenging noisy environments. An analysis of the learned probabilistic factors and resulting tracking performance is used to validate the approach. Results show that our approach can leverage the graph structure to report uncertainty about its estimates while significantly reducing the need for prior domain knowledge required by previous NBP methods. DNBP performs competitively in comparison to a traditional learning-based approach. Collectively, these results indicate that DNBP has the potential to be successfully applied to robotic perception tasks.

\section{Related Work}

\subsection{Belief Propagation}
Graphical models describe the joint distribution of a collection of random variables using a set of connected nodes comprising a graph structure. Connections between nodes in a graphical model represent probabilistic relationships between the corresponding random variables and are modeled by functions, referred to in this work as probabilistic factors. In the context of graphical models, inference refers to the process in which information about observed variables is used to derive the posterior distribution(s) of unobserved variables.

For inference on tree-structured graphs, belief propagation (BP) is proven to compute exact marginals~\cite{bp:PEARL1988143}. Loopy belief propagation (LBP)~\cite{lbp:MurphyWJ99} provides an approximation for inference with graphs containing loops and has demonstrated empirical success on a number of distinct tasks~\cite{lbp:MurphyWJ99,bpstereo:SunZS03,bpface:LeeASG08,bpdenoise:LanRHB06}. Both methods define a message passing scheme within the graph to infer the marginals of each unobserved variable (or hidden node in the graph). Due to their demand for exact summations within the marginalisation procedure, use of these methods is often infeasible in continuous domains where integrals become intractable.

In order to apply inference techniques such as BP and LBP, the parameters of a graphical model (e.g. the probabilistic factors) must be fully specified. Maximum likelihood estimation (MLE) has been shown to be an effective approach for learning the parameters of a graphical model from data~\cite{probperspbook:Murphy12,probgraphbook:koller2009}. Recently, Ping and Ihler~\cite{bpcrbm:PingI17} demonstrated the effectiveness of using MLE in conjunction with a matrix-based LBP algorithm for use on binary image denoising and image completion datasets. In contrast, this current study focuses on parameter learning for use with inference of continuous random variables.

\subsection{Nonparametric Belief Propagation}
For continuous spaces, such as six degrees-of-freedom object pose, exact integrals called for in BP and LBP become intractable and approximate methods for inference have been considered. Nonparametric belief propagation (NBP) methods \cite{isard2003pampas,nbp:SudderthIFW03}, have been proposed which represent the inferred marginal distributions using mixtures of Gaussians and define efficient message passing approximations for inference. Isard~\cite{isard2003pampas} demonstrated the effectiveness of PAMPAS using a set of synthetic visual datasets each modeled with hand-crafted factors. Sudderth et al. applied their NBP method successfully to a visual parts-based face localization task~\cite{nbp:SudderthIFW03} as well as a human hand tracking task~\cite{sudderth2004visual}. In both applications, NBP relied on factor models which were chosen based on task-level domain knowledge (e.g. valid configurations of human hands). Sigal et al.~\cite{looselimbed:SigalBRBI04} extended these NBP methods to human pose estimation and tracking using factors which were each trained separate from the inference algorithm using independent training objectives.

Ihler and McAllester~\cite{ihler2009particle} described a  conceptual theory of particle belief  propagation, where messages being sent to inform the marginal of a particular variable could be generated using a shared proposal distribution. This work emphasized the advantages of using a large number of particles to represent incoming messages, along with theoretical analysis. Following the work of Ihler and McAllester, Desingh et al.~\cite{bpposeest:DesinghLOJ19}, presented an efficient "pull" message passing algorithm (PMPNBP) which uses a weighted particle set to approximate messages between random variables. PMPNBP was shown to be effective on robot pose estimation tasks using hand-crafted factors. Using a similar approximation of belief propagation, Pavlasek et al. \cite{pavlasek2020parts} took a step toward neural network-based potential functions by introducing a pre-trained image segmentation network to the unary factors. Results from this study showed impressive performance for tool localization in clutter by pairing a nonparametric LBP approximation with neural network models.

An important limitation of these works is they assume the probabilistic factors expressed in the graph are provided as input or rely on domain knowledge to separately model and train each function. Recent work has explored the potential for neural networks to learn the parameters used by alternative inference techniques. Do and Arti{\`{e}}res~\cite{ncrf:DoA10} combined a conditional random field graphical model with multilayer neural networks using end-to-end training. Tompson et al.~\cite{cnngraph:TompsonJLB14} proposed a deep convolutional neural network that approximates marginal inference for a MRF graphical model using a network architecture designed to emulate the message passing of LBP. The resulting hybrid, which represented messages as discretized heat-maps, was trained in an end-to-end fashion and shown to outperform non-MRF based network architectures for human body pose estimation.
Xiong and Ruozzi~\cite{nnmrf:XiongR20} propose using MLE in conjunction with a Bethe free energy approximate inference algorithm to train neural networks modeling the probabilistic factors of an MRF graphical model.

This study sets out to explore the potential for a deep learning framework to be used within the inference process of PMPNBP such that the probabilistic factors may be learned in an end-to-end fashion. Although our current experiments are limited to simulated articulated bodies, we do not handcraft the learners specific to the domain of articulated tracking.

\subsection{Differentiable Bayes Filtering}
In the context of robot state estimation, many approaches have recently been proposed that incorporate neural networks with recursive inference algorithms. Haarnoja et al.~\cite{bpkf:HaarnojaALA16} introduced a differentiable Kalman filter for mobile robot state estimation. Jonshckowski and Brock~\cite{e2ehist:Jonschkowski-16-NIPS-WS} proposed a differentiable, histogram-based Bayes filter algorithm with end-to-end trained neural networks. Jonschkowski et al.~\cite{dpf:JonschkowskiRB18} and Karkus et al.~\cite{pfnet:KarkusHL18}, both proposed differentiable particle filter algorithms for modeling continuous state spaces. Lee et al.~\cite{multimodfilt:abs-2010-13021} investigate how multimodal sensor information may be fused for use in differentiable filtering algorithms. In each of these studies, the differentiable filtering algorithms were shown to outperform recurrent long short-term memory (LSTM) neural networks. 

In contrast to these methods, which model a single object body using variants of the Bayes filter, this work sets out to study the potential for PMPNBP to be used as an algorithmic prior for modeling multi-part articulated objects. The use of PMPNBP is motivated by its factorized inference process as an NBP method, which has been shown to contribute to improved performance over standard particle filter algorithms on articulated state estimation tasks~\cite{bpposeest:DesinghLOJ19,pavlasek2020parts}. 
Recently, this line of research on differentiable state estimation algorithms has extended into the planning domain~\cite{dancontrol:KarkusMHKLL19,dualsmc:WangL0ZD0T20,ghostplan:AndersonSPBL19}. Exploration of embedding DNBP within a differentiable planning system is left as future work.

\section{Belief Propagation}
Probabilistic graphical models, such as the MRF, describe probability distributions that can be inferred using BP. Consider an MRF model defined by the undirected graph $\mathcal{G}=\{\mathcal{V}, \mathcal{E}\}$, where $\mathcal{V}$ denotes a set of nodes and $\mathcal{E}$ denotes a set of edges. An example MRF model is shown in \cref{fig:pendulum_model}. Each node in $\mathcal{V}$ represents an observed or unobserved random variable while each edge in $\mathcal{E}$ represents a pairwise relationship between two random variables in $\mathcal{V}$. In addition, the MRF formulation specifies the joint probability distribution for this collection of random variables as:
\begin{equation}
\label{eq:jointprob}
    p(\mathcal{X}, \mathcal{Y}) = \frac{1}{Z} \prod_{(s,d)\in \mathcal{E}}\psi_{s,d}(X_s, X_d) \prod_{d \in \mathcal{V}}\phi_{d}(X_d, Y_d)
\end{equation}
where $\mathcal{X} = \{X_d\,\vert\,d\in \mathcal{V}\}$ is the set of unobserved variables and $\mathcal{Y} = \{Y_d\,\vert\,d\in \mathcal{V}\}$ is the set of corresponding observed variables. The scalar $Z$ is a normalizing constant. For each edge, the function $\psi_{s,d}(\cdot)$ is the \textit{pairwise potential}, describing the compatibility of an assignment for neighboring variables $X_s$ and $X_d$. 
For each node, the function $\phi_{d}(\cdot)$ is the \textit{unary potential}, describing the compatibility of an assignment for unobserved variable $X_d$ with an assignment for corresponding observed variable $Y_d$.

Given the factorization of the joint distribution defined in \cref{eq:jointprob}, BP provides an algorithm for exact inference of the marginal posterior distributions, or beliefs $bel(X_d)$, for tree structured graphs~\cite{bp:PEARL1988143}. BP defines a message passing scheme for calculation of the beliefs as follows:
\begin{equation}
\label{eq:bpbelief}
    bel_{d}(X_d)\propto \phi_d(X_d,Y_d)\prod_{s\in\rho(d)} m_{s\rightarrow d}(X_d)
\end{equation}
where each message is defined:
\begin{align}
\label{eq:bpmsg}
    m_{s\rightarrow d}(X_d)=\int_{X_s} & \phi_s(X_s,Y_s) \psi_{s,d}(X_s,X_d)\nonumber\\
    &\times \prod_{u\in\rho(s)\setminus d} m_{u\rightarrow s}(X_s) \;\;dX_s
\end{align}
where the term $\rho(s)$ denotes the set of neighboring nodes of $s$. Performing inference of random variables in continuous space causes the integral in Equation (\ref{eq:bpmsg}) to become intractable. This motivates the use of efficient algorithms that approximate the message passing scheme of Equations (\ref{eq:bpbelief})-(\ref{eq:bpmsg}).

\subsection{Nonparametric Belief Propagation}
\label{sec:pmpnbp}
In nonparametric belief propagation (NBP)~\cite{nbp:SudderthIFW03}, an iterative message passing algorithm is presented to infer the belief of random variables in continuous spaces. NBP uses Gaussian mixtures as a nonparametric representation of the beliefs and messages. A computationally expensive message generation process is used by NBP as an approximation of the integral in \cref{eq:bpmsg}. For complete details, we refer the reader to Sudderth et al.~\cite{nbp:SudderthIFW03}.

\subsection{Pull Message Passing Nonparametric Belief Propagation}
PMPNBP~\cite{bpposeest:DesinghLOJ19} avoids the expensive  message generation of NBP by approximating \cref{eq:bpmsg} with a ``pull'' strategy. The messages and beliefs of PMPNBP are represented by weighted and uniformly weighted particle sets of size $M$ and $T$ respectively:
\begin{align}
    bel^{n}_{d} &= \set{(\mu_d^{(i)})}_{i=1}^{T}\label{eq:pmpnbpbel}\\
    m^{n}_{s\rightarrow d} &= \set{(w_{sd}^{(i)},\mu_{sd}^{(i)})}_{i=1}^{M}
\end{align}
PMPNBP generates message, $msg^n_{s\rightarrow d}$, outgoing from $s$ to $d$, by first sampling $M$ independent samples from $bel_d(X_d)$ then reweighting and resampling from this set. In the following section, a differentiable variant of the PMPNBP algorithm is introduced that enables use of end-to-end training.

\section{Differentiable Nonparametric Belief Propagation}
\label{sec:dnbp}
We propose a differentiable version of PMPNBP, resulting in a differentiable nonparametric belief propagation (DNBP) method. The DNBP method leverages the strengths of both PMPNBP and neural networks: it can efficiently approximate the marginal posterior distributions encoded in an MRF, while learning the necessary potential functions in an end-to-end manner. Building on the description of PMPNBP~\cite{bpposeest:DesinghLOJ19}, this section describes the components of DNBP that enable end-to-end training of the potential functions.

\subsection{Unary Potential Functions}
According to the factorization encoded by the MRF formulation (\cref{eq:jointprob}), each unobserved variable $X_s$, for $s\in \mathcal{V}$, is related to a corresponding observed variable $Y_s$ by the unary potential function $\phi_s(X_s, Y_s)$. This function must be pointwise computable for an arbitrary setting of $X_s=x_s$ and $Y_s=y_s$. DNBP models each unary function with a feedforward neural network. The unary potential for a particle, $x_s$, given a particular observed image, $y_s$, is
\begin{align}
  \unary{s}{x_s}{y_s} &= l_s(x_s\oplus f_s(y_s))
\end{align}
where $f_s$ is a convolutional neural network, $l_s$ is a fully connected neural network, and the symbol $\oplus$ denotes concatenation of feature vectors. Details of network architectures are given in \cref{tab:netparams}.

\subsection{Pairwise Potential Functions}
This work considers MRF models limited to pairwise clique potentials. For any pair of hidden variables, $X_s,X_d$, which are connected by an edge in $\mathcal{E}$, there is a pairwise potential function, $\pairwise{s}{d}{}{}$, representing the probabilistic relationship between the two variables. DNBP models each pairwise potential using a pair of feedforward, fully connected neural networks, $\pairwise{s}{d}{}{}=\{\psi_{s,d}^{\rho}(\cdot), \psi_{s,d}^{\sim}(\cdot)\}$. The pairwise \textit{density} network, $\psi_{s,d}^{\rho}(\cdot)$, evaluates the unnormalized potential pointwise for a pair of particles. The pairwise \textit{sampling} network, $\psi_{s,d}^{\sim}(\cdot)$, is used to form samples of node $s$ conditioned on node $d$ and vice versa.

\subsection{Particle Diffusion}
DNBP uses a learned particle diffusion model for each hidden variable in place of the Gaussian diffusion models used by PMPNBP. Each diffusion process is modeled by DNBP as a distinct feedforward neural network, $\tau^{\sim}_{s}(\cdot)$ for $s\in\mathcal{V}$. These networks are used during the message passing algorithm to form samples of node $d$ at message iteration $t+1$ conditioned on a sample from iteration $t$.

\subsection{Particle Resampling}
PMPNBP performs a weighted resampling of belief particles as its final operation of the belief update algorithm. This operation is similar to the resampling performed in a standard particle filter and is non-differentiable~\cite{pfnet:KarkusHL18,dpf:JonschkowskiRB18,diffresamp:abs-2004-11938}. From the non-differentiable resampling, it follows that PMPNBP's belief update algorithm is non-differentiable. DNBP addresses the non-differentiability of the belief update algorithm by relocating the resampling and diffusion operations to the beginning of the message update algorithm (see Appendix \ref{appendix:alg_pseudo} for pseudo code). With this modification, the belief update returns a weighted set of particles approximating the marginal beliefs. The resulting belief density estimate is differentiable up to the beginning of the message update, when particles from the previous iteration were resampled.

Taken together with the description of PMPNBP\cite{bpposeest:DesinghLOJ19}, the components A-D. discussed in this section result in a differentiable, nonparametric approximation of belief propagation which can be used as an algorithmic prior.

\begin{table}[t]
\centering
\begin{tabular}{|c||l|l|l|l|}
\hline

Network              & \multicolumn{4}{l|}{Unit Layers}\\
\hline

$f_s$    & \multicolumn{4}{l|}{5 x [conv(3x3, 10, stride=2, ReLU), maxpool(2x2, 2)]}\\ 
\hline

$l_s$    & \multicolumn{4}{l|}{2 x fc(64, ReLU), fc(1, Sigmoid scaled to [0.005, 1])}\\ 
\hline

$\psi_{s,d}^{\rho}$    & \multicolumn{4}{l|}{4 x fc(32, ReLU), fc(1, Sigmoid scaled to [0.005, 1])}\\ 
\hline

$\psi_{s,d}^{\sim}$    & \multicolumn{4}{l|}{2 x fc(64, ReLU), fc(2)}\\ 
\hline

$\tau^{\sim}_{s}$    & \multicolumn{4}{l|}{2 x fc(64, ReLU), fc(2)}\\ 
\hline
\end{tabular}
\caption{Network parameters of learned DNBP potential functions. Note $s,d\in \mathcal{V}$, and $(s,d)\in \mathcal{E}$. Unary potentials: $l_s(f_s(\cdot))$. Pairwise potentials: $\{\psi_{s,d}^{\rho}, \psi_{s,d}^{\sim}\}$. Particle diffusion: $\tau^{\sim}_{s}$.}
\label{tab:netparams}
\end{table}

\subsection{Supervised Training}
DNBP's training approach is inspired by the work of Jonschkowski et al.~\cite{dpf:JonschkowskiRB18} with modifications to enable learning the potential functions distinct to DNBP. During training, DNBP uses a set of observation sequences, and a corresponding set of ground truth sequences. There is one observation sequence for each observed variable in $\mathcal{G}$ and one ground truth sequence for each unobserved variable in $\mathcal{G}$. Using the observation sequences, DNBP estimates belief of each unobserved variable at each sequence step. Then, by maximizing estimated belief at the ground truth label of each unobserved variable, DNBP learns its network parameters by maximum likelihood estimation. Next, the objective function and network gradients distinct to DNBP training are discussed.

\subsubsection{Objective Function}
Note that the belief weight, $w_d^{t,(i)}$, of particle $i$ is proportional to the product of \textit{component} weights, $w^{t,(i)}_{unary_d}\times w^{t,(i)}_{unary_s}\times w^{t,(i)}_{neigh_s}$, where $s$ is the neighbor of node $d$ from which particle $i$ originated. Since each of these component weights is produced by a separate potential network (either $\phi_d$, $\psi_{s,d}^{\sim}$, or $\psi_{s,d}^{\rho}$ respectively), direct optimization of the belief density will lead to interdependence of the potential network gradients during training. In the context of DNBP, interdependence between different potential functions is inconsistent with the factorization given in $\mathcal{G}$. Tompson et al.~\cite{cnngraph:TompsonJLB14} describe a similar phenomenon they refer to as gradient coupling which was addressed by expressing a product of features in log-space which `decouples' the gradients.

To avoid interdependence between potential functions during training, we consider the \textit{partial}-belief densities which are defined for each node $d\in \mathcal{V}$ as mixtures of Gaussian density functions:
\begin{align}
    &\overline{bel}^{t}_{d,unary_d}(X_d) = \sum_{i=1}^N w^{t,(i)}_{unary_d}\cdot \mathcal{N}(X_d; \mu_d^{(i)}, \Sigma) \label{eq:pseudobellik} \\
    &\overline{bel}^{t}_{d,unary_{\rho(d)}}(X_d) = \sum_{i=1}^N w^{t,(i)}_{unary_s}\cdot \mathcal{N}(X_d; \mu_d^{(i)}, \Sigma) \label{eq:pseudobelunary} \\
    &\overline{bel}^{t}_{d,neigh_{\rho(d)}}(X_d) = \sum_{i=1}^N w^{t,(i)}_{neigh_s}\cdot \mathcal{N}(X_d; \mu_d^{(i)}, \Sigma) \label{eq:pseudobelneigh}
\end{align}
Using these definitions, direct interaction between the potential networks' gradients is avoided by maximizing the product of partial-beliefs at the ground truth of each node in log space. The product of partial-beliefs is defined:
\begin{align}
    \overline{bel}^t_d(X_d) = \;&\overline{bel}^{t}_{d,unary_d}(X_d)\times \overline{bel}^{t}_{d,unary_{\rho(d)}}(X_d) \nonumber\\
    &\times \overline{bel}^{t}_{d,neigh_{\rho(d)}}(X_d)
\label{eq:pseudbel}
\end{align} 

With this construction, DNBP defines a loss function for each hidden node $d\in\mathcal{G}$ as:
\begin{align}
    L^t_d = -\log(\overline{bel}^t_d(x^{t,*}_d))
\label{eq:loss}
\end{align}
where $x^{t,*}_d$ denotes the ground truth label for node $d$ at sequence step $t$. At each sequence step during training, DNBP iterates through the nodes of the graph, updating each node's incoming messages and belief followed by a single optimization step of \cref{eq:loss} using stochastic gradient descent.

\subsubsection{Unary Potentials}
During training of DNBP, only those gradients derived from the belief update of each node are used to update the corresponding node's unary potential network parameters. Any gradients derived from the outgoing messages of a particular node are manually stopped from propagating to that node's unary network. This is done to avoid confounding the objective functions of neighboring nodes, which each rely on the others' unary network during message passing. This approach can be implemented with standard deep learning frameworks by dynamically stopping the parameter update of each unary network depending on where in the algorithm its forward pass was registered.

\subsubsection{Pairwise Density Potentials}
To speed up and stabilize the training of pairwise density potential networks, the following substitution is made during training. While calculating $w_{neigh}^{(i)}$ for outgoing message $i$ from node $s$ to $d$, the summation over incoming messages from $u\in \rho(s)$ to $s$ is replaced by a single evaluation of 
\begin{equation}
W_u^{(i)}=\pairwise{s}{d}{x^{t,*}_s}{\prt{s}{d}{i}}
\end{equation}
where $x^{t,*}_s$ is the ground truth label of sender node $s$. This change improves inference time and reduces memory demands by removing a summation over $M$ particles while also providing more stable training feedback to the network. This substitution is removed at test time after training is complete.

\section{Experimental Setup}
The capability of DNBP is demonstrated on two challenging articulated tracking tasks. The performance of DNBP is compared with a baseline LSTM neural network \cite{lstm:HochreiterS97} on both tasks. LSTM was chosen as a baseline because of its established performance on visual tracking tasks~\cite{deepvistrack:LI2018323}, thus LSTM serves as a positive control in these experiments. In the first task, both DNBP and LSTM attempt to track the position of each joint of a simulated double pendulum, as shown in \cref{fig:pendulum_example}, that swings under the effect of gravity. The second task involves tracking a simulated `spider' structure containing both prismatic and revolute joints. In each task, articulated state is represented as a collection of 2-dimensional keypoints, where the position of each joint corresponds to a single keypoint. DNBP models the kinematic structure of each task using a separate MRF model, in which each unobserved variable corresponds to the position a particular keypoint and each observed variable corresponds to an observed RGB image. In this model, each edge represents the articulation constraint between a pair of object parts. 

To emulate environments where occlusion is common, simulated clutter in the form of static and dynamic geometric shapes are rendered into the image sequences. Simulated static clutter remains stationary throughout image sequences while dynamic clutter exhibits translational and rotational motion. This clutter is challenging as the shapes are visually similar to the parts of the target structure and hence distracting to the unary potentials. \textit{Clutter ratio} is used to measure the amount of clutter in a single sequence. In this work, clutter ratio is defined as the ratio of pixels occluded by simulated clutter to the total number of image pixels and is averaged over the full sequence.

\subsection{Double Pendulum Tracking}
\label{sec:experiment_pendulum}

\begin{figure}[t!]
\centering
\begin{subfigure}[b]{0.5\columnwidth}
\centering
\includegraphics[width=0.89\columnwidth]{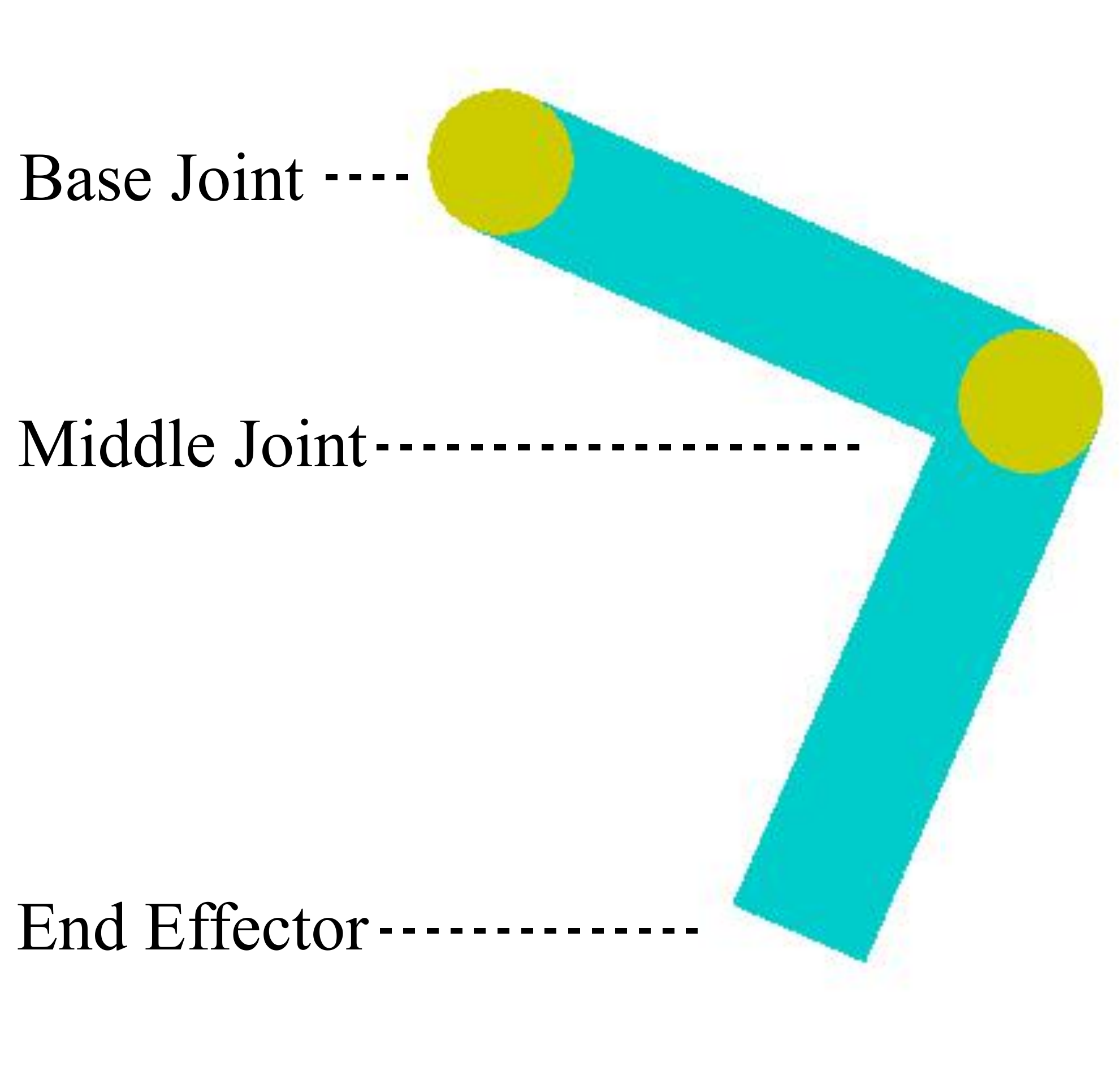}
\caption{}
\label{fig:pendulum_example}
\end{subfigure}\hfill%
\begin{subfigure}[b]{0.5\columnwidth}
\centering
\includegraphics[width=0.545\columnwidth]{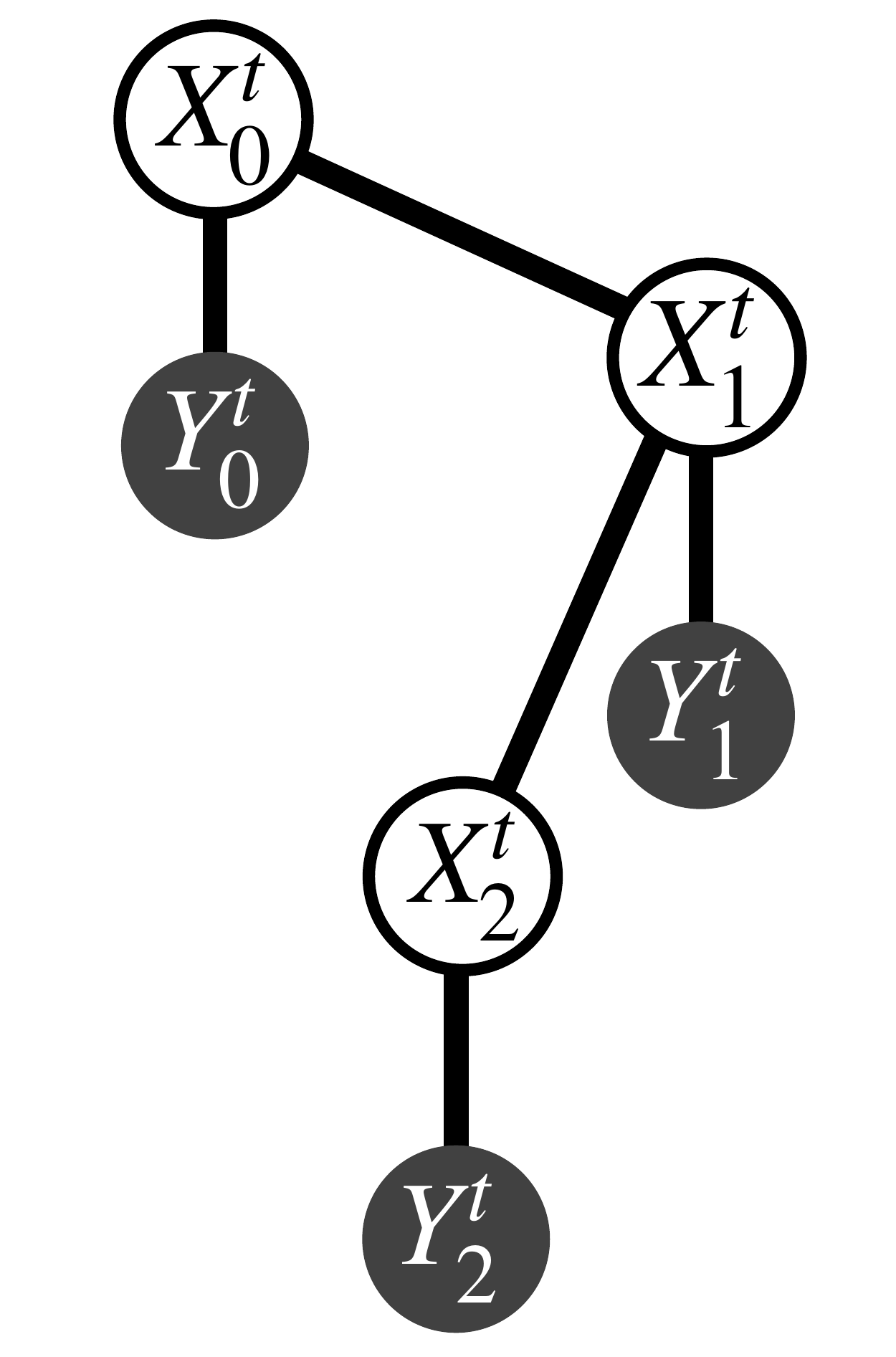}
\caption{}
\label{fig:pendulum_model}
\end{subfigure}
\caption{a) Geometry and example configuration of the double pendulum. b) Graphical model used by DNBP for the double pendulum task.}
\end{figure}

The double pendulum task was chosen to compare the tracking performance of each method using an articulated structure exhibiting chaotic motion. The double pendulum structure consists of two revolute joints connected to two rigid-body links in series. A modified version of the OpenAI Gym~\cite{openai:1606.01540} Acrobot environment is used to generate a simulated double pendulum dataset. The default Acrobot environment time step parameter was set to $0.08t$ and both links' lengths were set to $0.8m$. In each simulated sequence the joint states are randomly initialized in the range of $[0,2\pi]$ and each joint velocity to $0$. No control signal is used, thereby simulating motion affected only by the initial configuration and gravity.

An example of the double pendulum simulated by this environment is shown in \cref{fig:pendulum_example}. The pose of the double pendulum is modeled by the position of its two revolute joints, rendered as yellow circles, and one end effector. The corresponding graphical model used by DNBP is shown in \cref{fig:pendulum_model}. Hidden variable $X_0$ corresponds to the double pendulum's base joint keypoint, $X_1$ to the middle joint keypoint and $X_2$ to the end effector keypoint. The observed variables, $Y_0,Y_1$, and $Y_2$, represent the observed images. The goal of the double pendulum tracking task is to predict the position of each keypoint on the double pendulum for each frame of a given input image sequence. A successful perception system must handle the chaotic motion inherent to the double pendulum as well as cases of occlusion. 

Both DNBP and the LSTM baseline are provided a training set used to inform their predictions, a validation set to tune training parameters and are evaluated on a held-out test set. The training set consists of $1,024$ total sequences with $20$ frames per sequence while the validation set consists of $150$ total sequences with $20$ frames per sequence. Both training and validation sequences are split evenly among three bins of clutter ratio: none, $0$ to $0.04$ and $0.04$ to $0.1$. Of the training and validation sequences with any amount of clutter, half contain static clutter and the other half contain dynamic clutter. The held-out test set is evenly split among clutter ratio deciles from $0$ to $0.95$, thus contains a shift in distribution from the training set, which was limited to clutter ratios below $0.1$. Each decile contains $50$ sequences with $100$ frames per sequence. For test sequences with any amount of clutter, half contain static clutter and the other half contain dynamic clutter. For additional details on the clutter generation, see \cref{appendix:pendulum_clutter}.

\subsection{Articulated Spider Tracking}
\label{sec:experiment_spider}

\begin{figure}[t!]
\centering
\begin{subfigure}[b]{0.5\columnwidth}
\centering
\includegraphics[width=0.65\columnwidth,keepaspectratio]{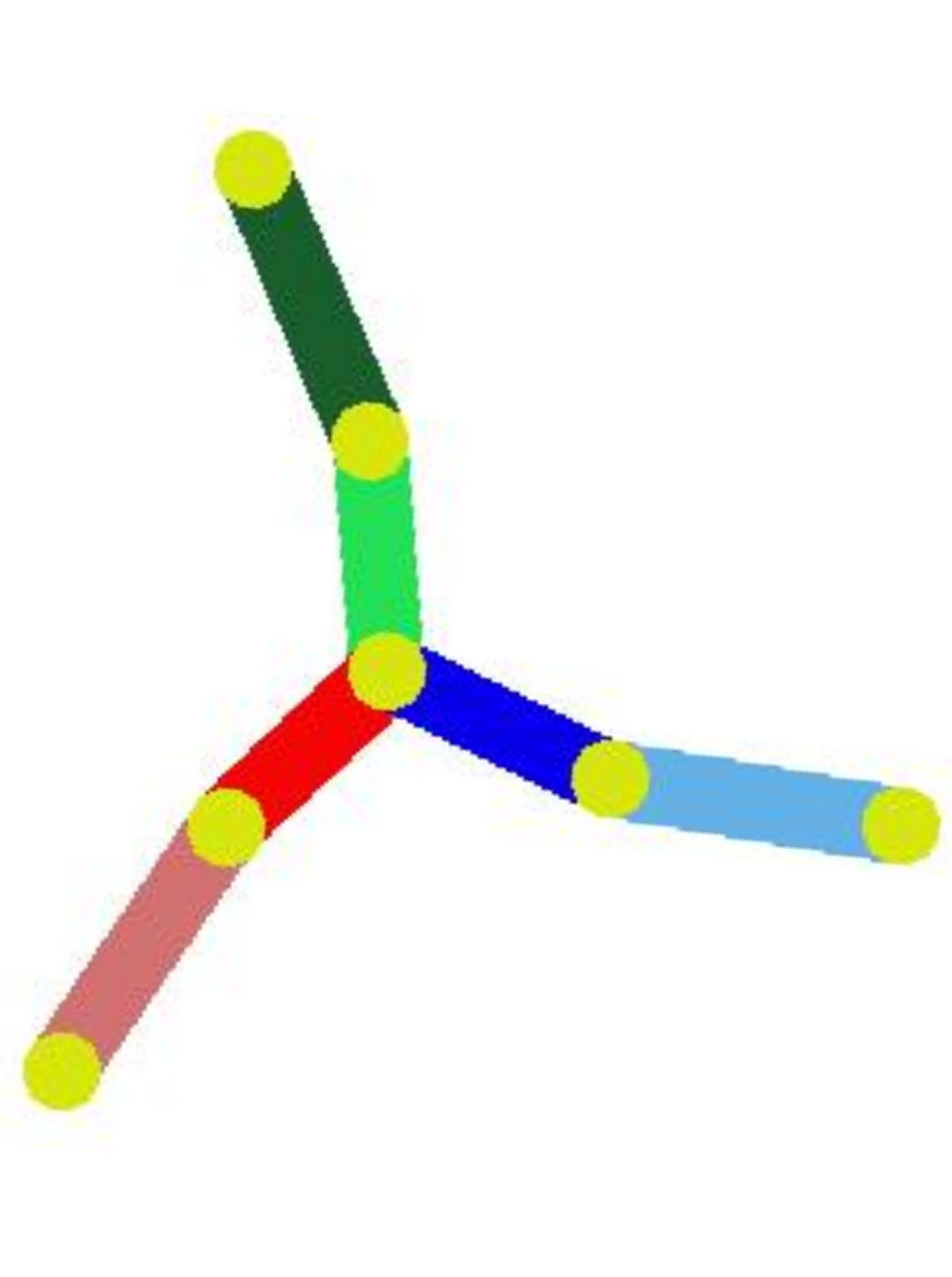}
\caption{}
\label{fig:spider_example}
\end{subfigure}\hfill%
\begin{subfigure}[b]{0.5\columnwidth}
\centering
\includegraphics[width=0.65\columnwidth,keepaspectratio]{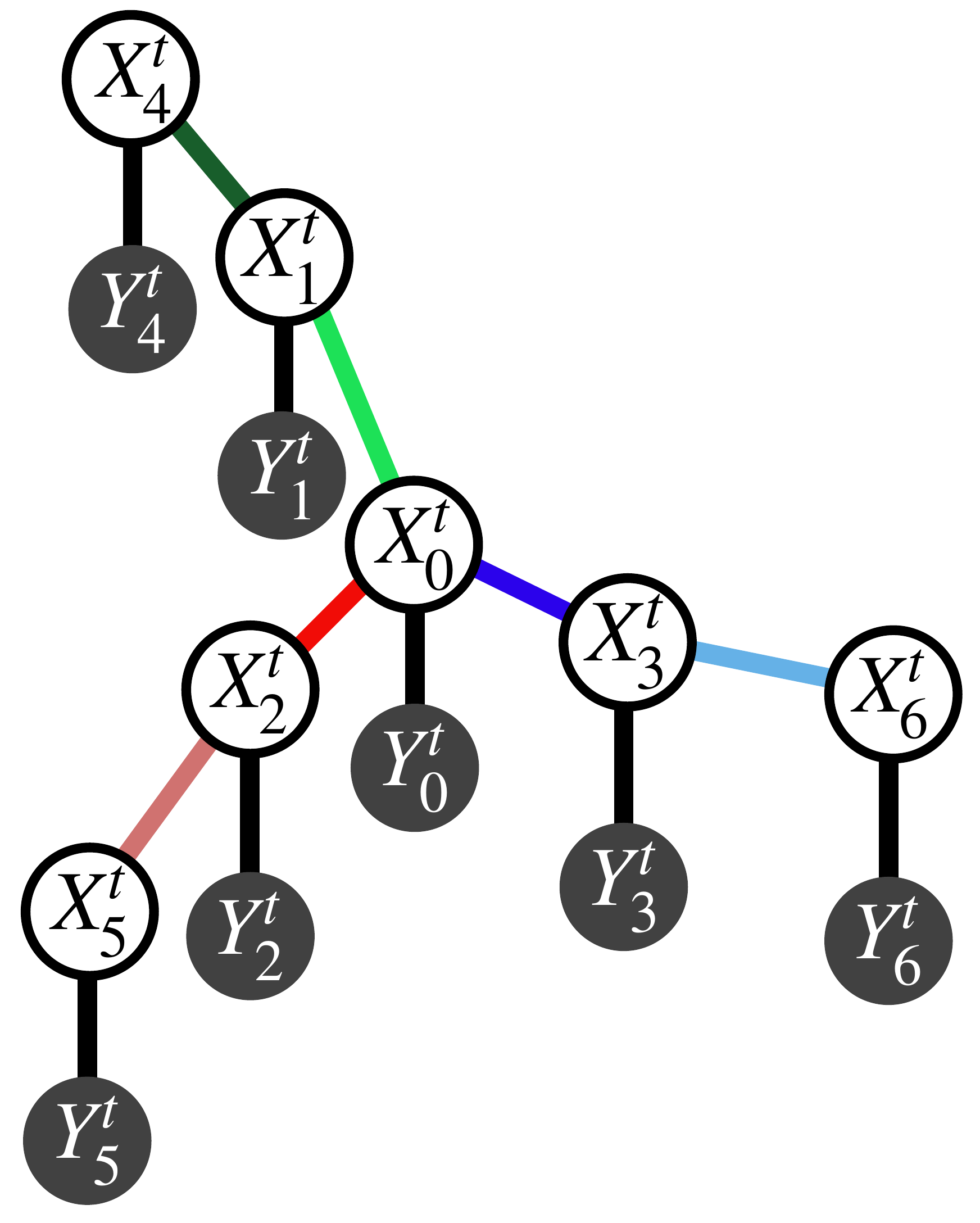}
\caption{}
\label{fig:spider_model}
\end{subfigure}
\caption{a) Geometry and an example configuration of the spider structure. b) Graphical model used by DNBP for the spider task.}
\end{figure}

The `spider' task was chosen to further characterize DNBP's performance using a structure with added articulations and a larger graphical model. The structure is comprised of three revolute-prismatic joints, three purely revolute joints, and six rigid-body links. An example of the spider is shown in \cref{fig:spider_example}, in which the joints are rendered as yellow circles and the rigid-body links are rendered as coloured rectangles. Additional details of this structure and how it is simulated are provided in \cref{appendix:spider_model}. The corresponding graphical model used by DNBP is shown in \cref{fig:spider_model}, in which each hidden variable corresponds to the keypoint location of a single joint and each observed variable corresponds to the input image. The goal of this task is to predict the position of each keypoint on the spider for each frame of a given input image sequence. Unlike the double pendulum, which contained a stationary base joint, the spider is not tethered to any position and can move freely throughout the image. The increased degrees of freedom and number of articulations in the spider task is intended to evaluate the adaptability of DNBP using a more complicated articulated structure.

The training, validation and test set for this task follow the same respective distributions of clutter as were used in the double pendulum datasets. The training set consists of $2,048$ total sequences and the validation set consists of $300$ sequences. The training and validation sequences are split evenly among five bins of clutter ratio: none, $0$ to $0.04$ and $0.04$ to $0.1$, $0.1$ to $0.2$ and $0.2$ to $0.3$. There are $20$ frames per sequence in each of the spider datasets. Additional details of the spider clutter generation are provided in~\cref{appendix:spider_clutter}.

\subsection{LSTM Baseline}
For both tasks, DNBP is compared to an LSTM recurrent neural network~\cite{lstm:HochreiterS97}. The LSTM baseline for these experiments was modeled on the architecture used by Jonschkowski et al.~\cite{dpf:JonschkowskiRB18}. The baseline architecture is composed of an encoder network, a $2$-layer LSTM, and a separate decoder network for each keypoint ($3$ decoders for double pendulum, $7$ decoders for spider). The encoder network is composed of 5x[conv(3x3, $N_{\text{Enc}_1}$, stride 1, ReLU), maxpool(2x2, stride 2)] followed by a single fc($N_{\text{Enc}_2}$, ReLU) layer. The 2-layer LSTM is attached to the encoder output and uses $N_{\text{Enc}_2}$ features. Finally, the LSTM output is fed to a set of decoder networks, each composed of [fc(64, ReLU), fc(32, ReLU), fc(2, ReLU)]. The LSTM is trained using average Euclidean distance and cutoff once validation loss stops improving. For both tracking tasks, LSTM is trained to minimize the average Euclidean error using window size of $20$ frames.

LSTM feature dimensions were chosen to keep the total number of parameters comparable between LSTM and DNBP. On the double pendulum task, $N_{\text{Enc}_1}=32$ and $N_{\text{Enc}_2}=46$, resulting in $89,460$ trainable parameters for the LSTM baseline, while DNBP uses $76,666$ parameters. On the spider task, $N_{\text{Enc}_1}=48$ and $N_{\text{Enc}_2}=64$, resulting in $198,318$ trainable parameters for the LSTM baseline, while DNBP uses $179,799$ parameters.

\subsection{Implementation Details}
\label{sec:implementation_details}
All train and test images are scaled to $128\times 128$ pixels. Ground truth keypoint locations are $2$-dimensional, continuous valued coordinates scaled to range of $[-1,+1]$. During training, additive Gaussian noise is applied to each image with $\sigma=20$. Both DNBP and LSTM are optimized using Adam~\cite{adamopt:KingmaB14} with a batch size of $6$ and trained until convergence of the validation loss. DNBP is trained using $100$ particles per message and tested using $200$ particles per message. The maximum weighted particle from each marginal belief set of DNBP is used during evaluation for comparison with the ground truth. Output from the LSTM is compared directly with the ground truth during evaluation.

To ensure independence from spatial location, the pairwise density, pairwise sampling and diffusion sampling processes are defined over the space of transformations between variables. Each pairwise density network, $\psi_{s,d}^{\rho}$, takes as input the difference between pairs of particles, $x_s-x_d$. The pairwise sampling networks, $\psi_{s,d}^{\sim}$, take a random sample of Gaussian noise as input and generate conditional samples using the following rule:
\begin{align}
    \epsilon&\sim\mathcal{N}(0,1)\\
    x_{s\vert d}&= x_d + \psi_{s,d}^{\sim}(\epsilon)
\end{align}
where $x_{s\vert d}$ is the sample of variable $X_s$ conditioned on neighboring sample $x_d$ and where $\epsilon$ is a noise vector sampled from a zero-mean, unit variance multivariate Gaussian distribution with $dim(\epsilon)=64$. Similarly, for sampling in the opposite conditioning direction (node $d$ conditioned on $s$), the same network is used but the sampled translation is negated. The diffusion process also is defined over the space of transformations but from $d^t$ to $d^{t+1}$. That is,
\begin{align}
    \epsilon&\sim\mathcal{N}(0,1)\\
    x^{t+1}_d&=x^{t}_d + \tau^{\sim}_{d}(\epsilon)
\end{align}
where $\epsilon$ is again a normal distribution with $dim(\epsilon)=64$.

\newcommand{\norm}[1]{\left\lVert#1\right\rVert}

\subsection{Performance metrics}
As a quantitative measure of tracking error, average Euclidean error is used. Given a set of predicted and true keypoints for a collection of images, the Euclidean distance between each predicted and true keypoint is calculated then averaged over every pair. 

Discrete entropy~\cite{entropy:Shannon48} is used as a quantitative measure of uncertainty estimated by DNBP. Given a set of marginal belief estimates produced by DNBP, the uncertainty measure is calculated as follows: First a weighted resampling operation is applied to the belief sets, resulting in a set of unweighted particles for each marginal. The particles are then binned to form a $40$x$40$ discrete histogram (equally divided over pixel space) for each marginal. Finally, the discrete entropy of each normalized histogram is calculated to produce a quantitative estimate of the marginal uncertainty associated with each keypoint estimate from DNBP. Bin size was chosen manually to ensure tractable calculation.

For qualitative analysis of the uncertainty estimated by DNBP, samples from an approximation of the joint posterior distribution are formed using a sequential Monte Carlo sampling approach~\cite{smc:NaessethLS14}. These samples are formed using the following iterative procedure: First, $N$ particles are sampled from the marginal belief of $X_0$. Next, iterate over each hidden variable (in the order defined by graph indices) and take $N$ samples from the corresponding marginal belief set which is reweighted by the pairwise potential, which can be pointwise computed using the $N$ partial samples of previous iterations. The output of this procedure is $N$ samples, each with a setting for all $|\mathcal{V}|$ keypoints. These samples are used only for qualitative evaluation of the uncertainty. Visualization of these samples are formed by plotting a rendered link between each pair of keypoint samples.
\section{Results}

\subsection{Double Pendulum Tracking}

\begin{figure}[t]
\centering
\includegraphics[width=0.95\columnwidth,keepaspectratio]{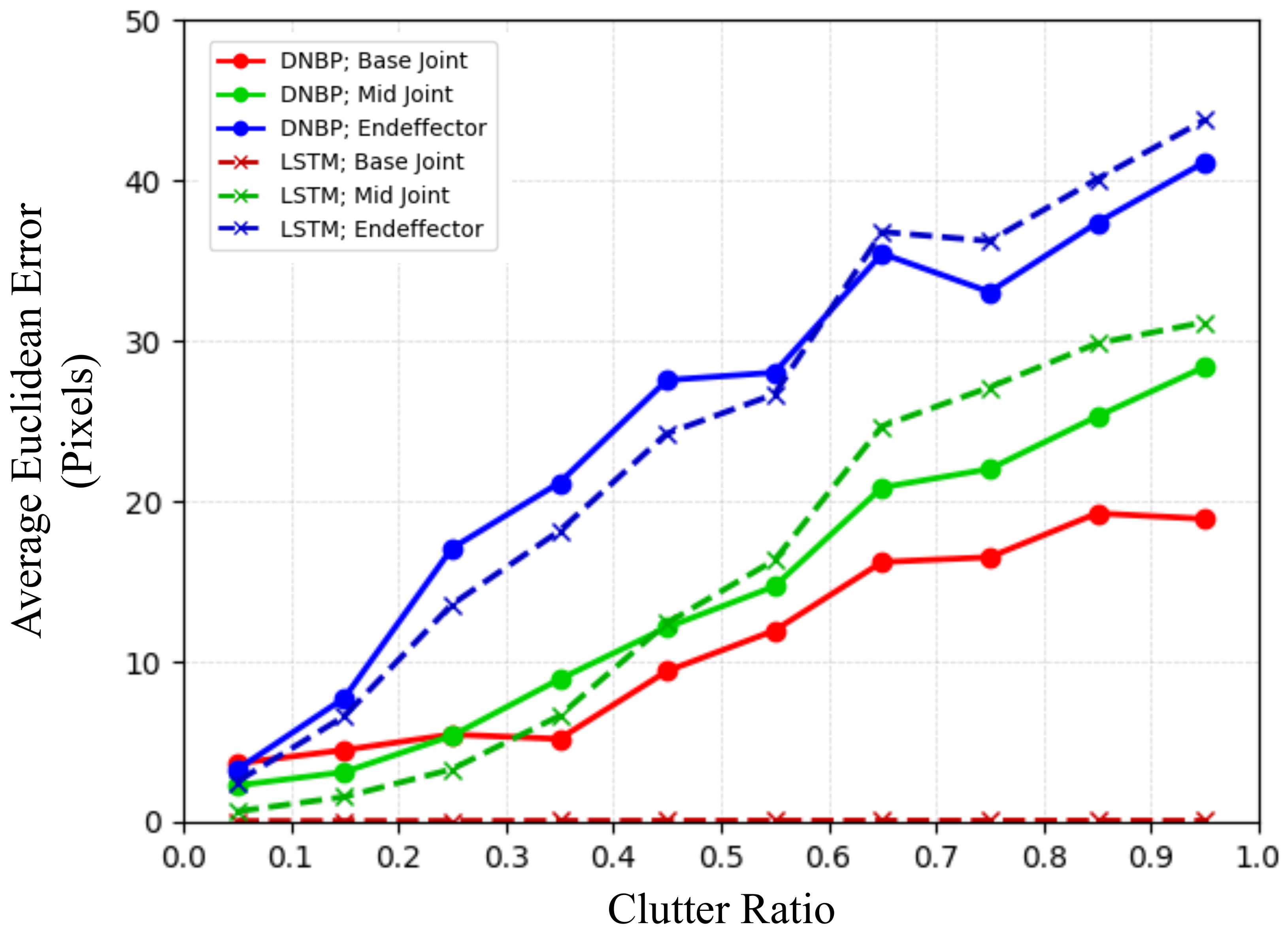}
\caption{Double pendulum keypoint tracking error. Average error of DNBP and LSTM predictions as a function of clutter ratio and keypoint type. Test sequences are binned as indicated by clutter ratio, which is calculated by fraction of pixels added as noise.}
\label{fig:pendulum_error}
\end{figure}

The performance of DNBP on the double pendulum tracking task was evaluated initially. As shown in \cref{fig:pendulum_error}, the keypoint tracking error of DNBP is directly compared to that of the LSTM baseline on the held-out test set for each keypoint type (base, middle and end effector) across the full range of clutter ratios. Results from this comparison show that DNBP's average keypoint tracking error is comparable to the LSTM's corresponding error for both the mid joint and end effector keypoints, independent of clutter ratio. For the base joint keypoint, which is stationary at the center position of every image, the LSTM was able to memorize the correct position. DNBP, which diffuses particles based on the message passing scheme, does not memorize the base joint position and registers a consistently larger error which increased with clutter ratio.

\begin{figure}[t!]
\centering
\includegraphics[width=1\linewidth]{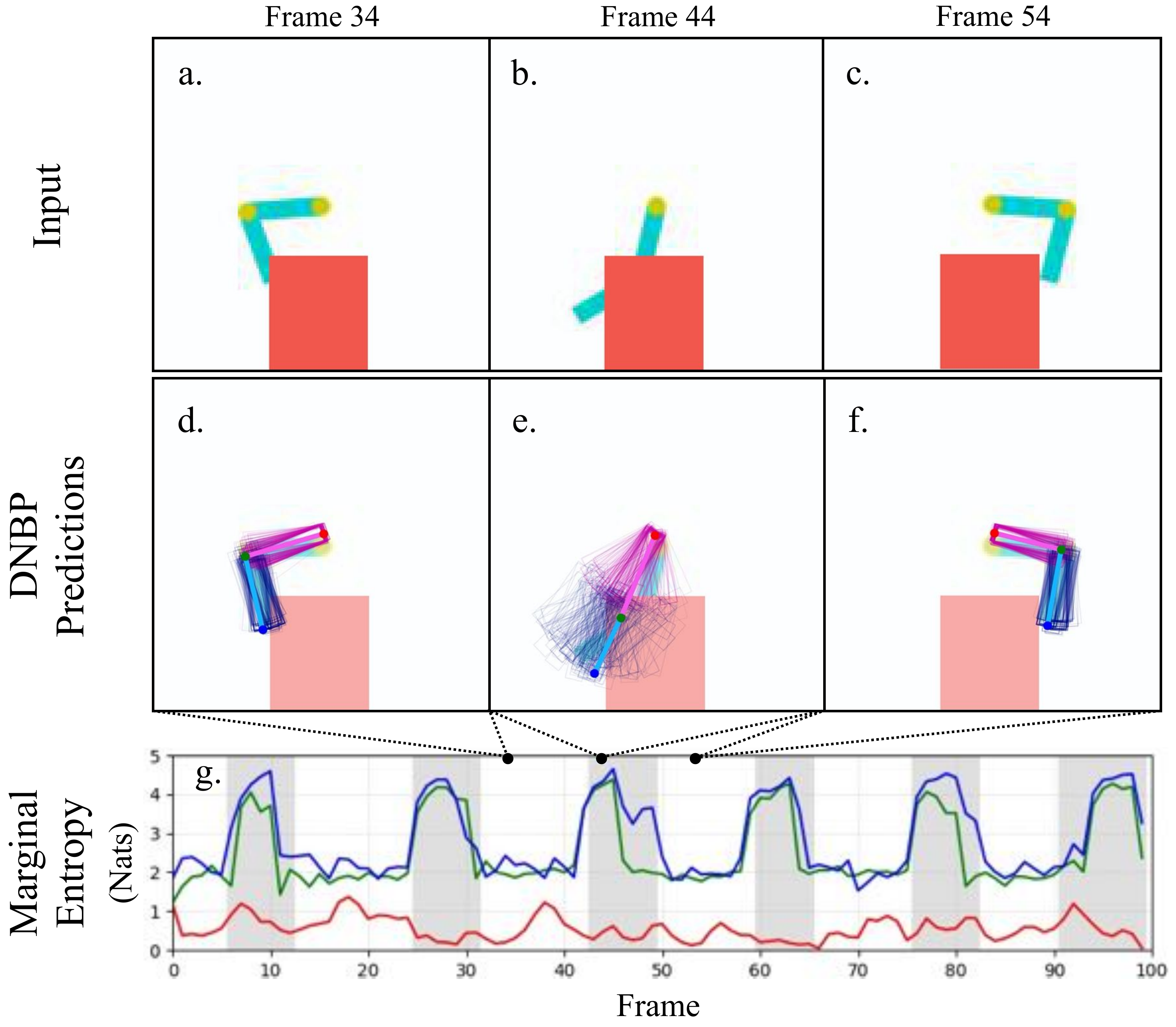}
\caption{Tracking of double pendulum by DNBP under partial occlusion. a,b,c: indicated input frames with occlusion drawn as orange block. d,e,f: corresponding output from DNBP overlaid on input image where final prediction is represented by red, green, and blue circles for the base, middle, and end-effector joints respectively. Uncertainty associated with predictions is shown qualitatively as samples from the joint distribution drawn as pendulum arms in pink and blue. g: marginal entropy for each keypoint across test sequence; base keypoint (red), middle keypoint (green), end-effector keypoint (blue). In g., sequence steps highlighted by gray correspond to images in which $>25\%$ of the pendulum's surface area is occluded.}
\label{fig:pendulum_occ}
\end{figure}

DNBP provides measures of uncertainty associated with its predictions, which are generated according to the algorithmic prior of belief propagation (PMPNBP). Next tested was the hypothesis that the DNBP model would generate increased uncertainty under conditions in which an occluding object is placed into the input images such that it covers portions of the double pendulum. This test was performed by rendering an occluding block onto a test sequence as shown in \cref{fig:pendulum_occ}a-c. Under optimal conditions, in which the pendulum is minimally occluded ($<25\%$ by surface area), the model's output indicates a low level of uncertainty (see \cref{fig:pendulum_occ}d,f,g.) for each keypoint and each frame. In contrast, under conditions in which the pendulum is occluded by the superimposed object, the model's output indicates relatively high levels of uncertainty precisely at frames in which the superimposed object occludes a portion ($>25\%$) of the double pendulum (see \cref{fig:pendulum_occ}e,g.). These results demonstrate that the estimate of uncertainty produced by DNBP can identify predictions which are unreliable.

\begin{figure}[t]
\centering
\includegraphics[width=0.9\linewidth]{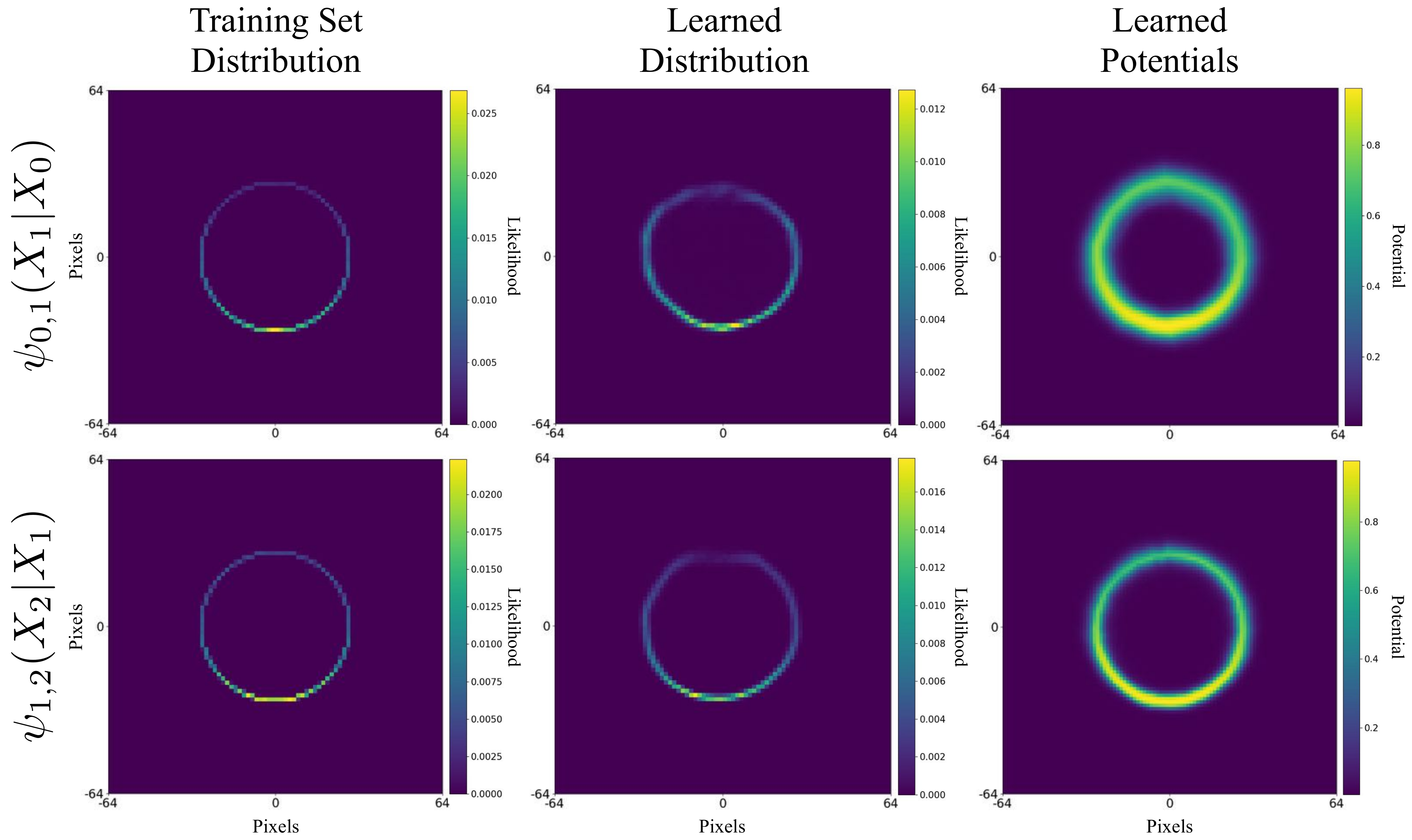}
\caption{Inspection of learned pairwise potentials from double pendulum tracking.}
\label{fig:learned_pairwise_pendulum}
\end{figure}

As further validation of DNBP, the learned pairwise potentials are inspected in \cref{fig:learned_pairwise_pendulum}. The normalized histogram of pairwise translations computed from the training set for $X_1-X_0$ (top) and $X_2-X_1$ (bottom) are shown in the left column of \cref{fig:learned_pairwise_pendulum}. The middle column shows the normalized histogram of samples from learned pairwise sampler networks, $\psi_{0,1}^{\sim}(\cdot)$ and $\psi_{1,2}^{\sim}(\cdot)$. Finally, the right column shows output from the learned pairwise density networks, $\psi_{0,1}^{\rho}(\cdot)$ and $\psi_{1,2}^{\rho}(\cdot)$,  generated with $100$x$100$ uniform samples across pairwise translation space. The qualitative similarity between each learned potential model and the corresponding true distribution of pairwise translations indicates that DNBP is successful in learning to model each pairwise potential factor. The circular pairwise relationships are explained by the fact that each pair of double pendulum keypoints is related by a revolute joint. The effect of simulated gravity in the double pendulum experiment can be observed by the bias of each pairwise potential in favor of the lower half of each plot as indicated by increased likelihood.

\subsection{Articulated Spider Tracking}

After having established the basic performance characteristics of DNBP on the relatively straightforward double pendulum task, we next set out to determine DNBP's capability for tracking more complex articulated structures. To this end, the 3-arm spider structure was used as a more challenging articulated pose tracking task. Details of this task are provided in \cref{sec:experiment_spider}, but crucially the spider contains both revolute and prismatic joints as well as distinct articulation constraints for the various joints. Similar to experiments on the double pendulum task, image complexity is further increased by adding static and dynamic geometric shapes (not part of the spider) superimposed onto the images.

\begin{figure}[t]
\centering
\includegraphics[width=0.95\columnwidth,keepaspectratio]{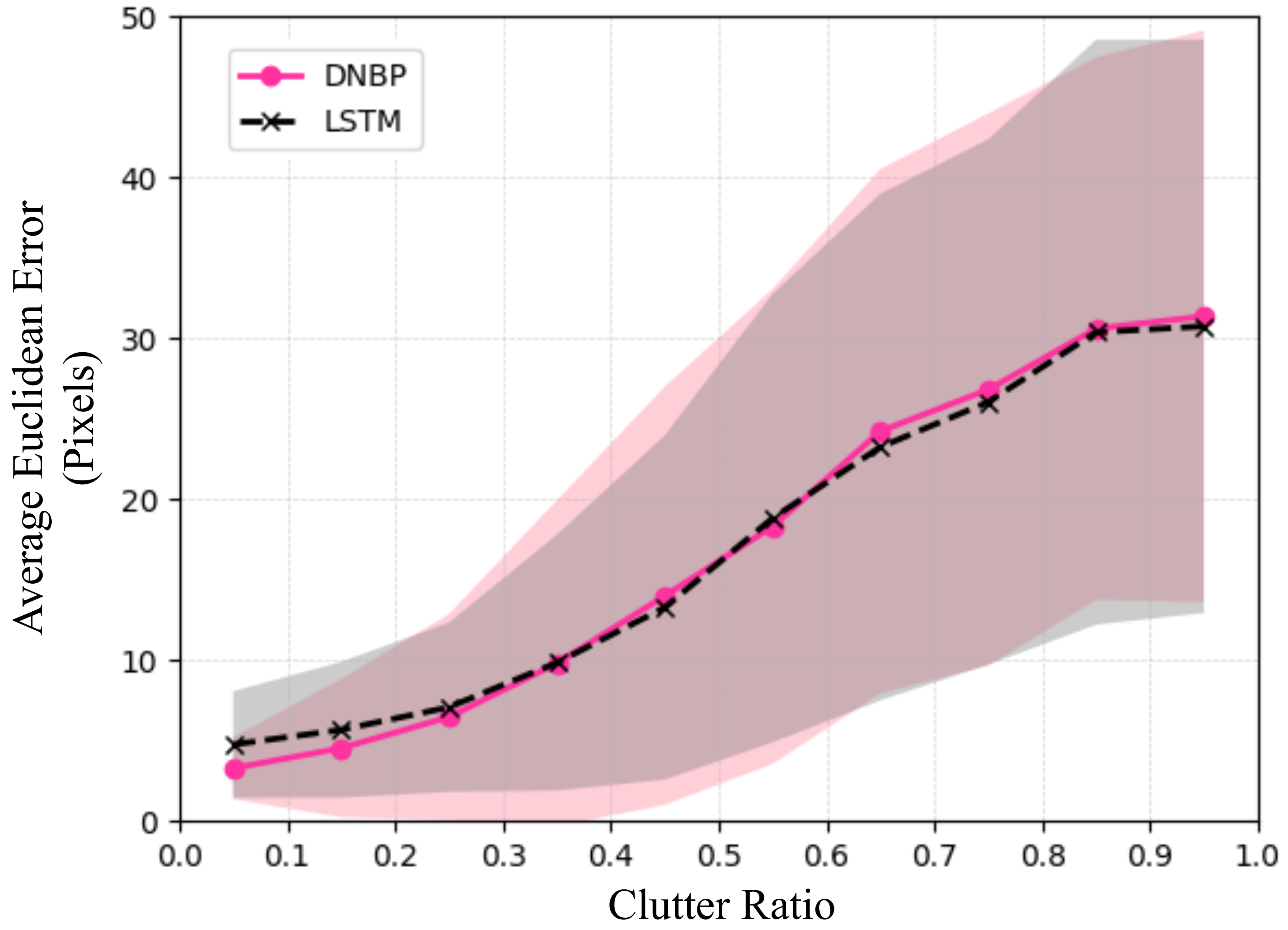} 
\caption{Average error as a function of clutter ratio and DNBP particle count. The mean value of the error in predicted position of each keypoint across the entire dataset is presented as average error. Test sequences are binned as indicated by clutter ratio, which is calculated by fraction of pixels added as noise. Particle count used in DNBP model varied as indicated. LSTM average error is shown as black dashed line and is independent of particle count.}
\label{fig:spider_error}
\end{figure}

As an initial evaluation, each model's performance is quantitatively assessed on the held-out test set of the articulated spider tracking task using the same approach as described for the double pendulum experiment by varying clutter ratio (\cref{fig:spider_error}). Similar to the results of the double pendulum experiment, average error on the spider task increases as a function of clutter ratio for both the LSTM and for DNBP. For clutter ratios between $0$ and $0.25$, average error for both models remains near $6$ pixels then increases consistently with clutter ratio, reaching above $30$ pixels of average error for clutter ratios above $0.85$. As in the case of the double pendulum experiment, these results demonstrate comparable performance between LSTM and DNBP on an articulated pose tracking task.

\begin{figure}[t!]
  \includegraphics[width=\columnwidth]{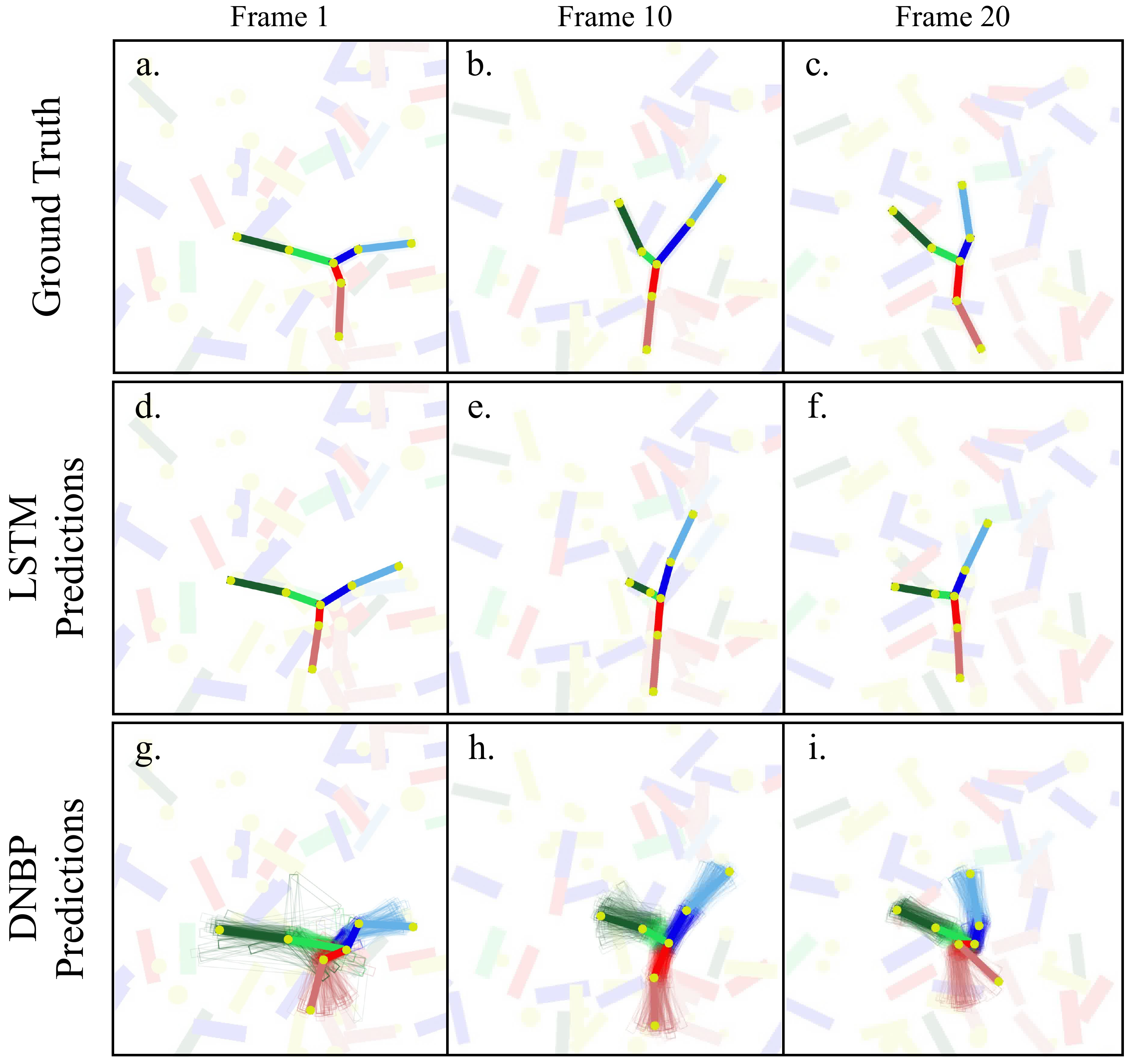}
  \caption{Comparison of articulated 'spider' tracking by LSTM and DNBP under cluttered conditions. a,b,c: ground truth pose at indicated frames. d,e,f: corresponding predictions from LSTM. g,h,i: predictions from DNBP. Predicted and ground truth keypoints shown as yellow circles with arms distinguished by color. Clutter shown here as faded shapes for illustration to highlight predictions.}
\label{fig:spidertrack}
\end{figure}

Next, a qualitative example of tracking performance under conditions of clutter is shown in \cref{fig:spidertrack}. In \cref{fig:spidertrack}(a-c), the ground truth spider pose is shown amidst distracting shapes across selected frames of a test sequence with clutter ratio of $0.25$. Pose predictions generated by LSTM are shown in \cref{fig:spidertrack}(d-f) and by DNBP in (g-i). Qualitative assessment of the images shown indicates that both the LSTM and DNBP place their spider predictions in the correct region of the image. Additionally, each model is shown to correctly predict the relative positions of the three spider arms. Over the sequence, both models track the prismatic and rotational motion of each keypoint, however appear to struggle with certain keypoint predictions. On frame $10$ (\cref{fig:spidertrack}b,e,h.), the LSTM incorrectly predicts the green arm is retracted near the spider's center while DNBP predicts the arm is extended with some degree of uncertainty. On frame $20$ (\cref{fig:spidertrack}c,f,i.), LSTM correctly predicts the red arm as extended while DNBP incorrectly predicts the red arm as being partially retracted.

\begin{figure}[t]
\centering
\includegraphics[width=0.9\linewidth]{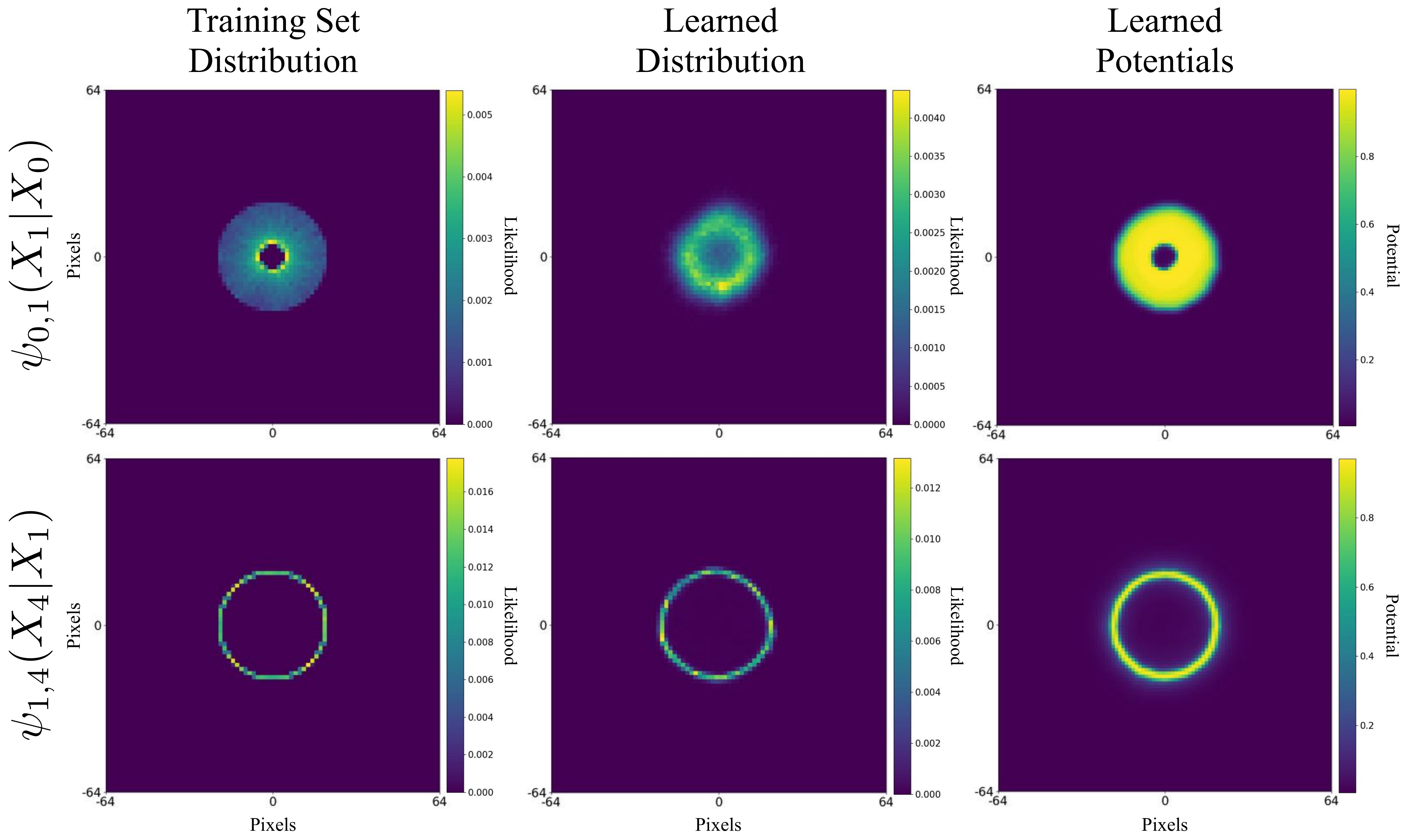}
\caption{Inspection of DNBP's learned pairwise potentials from spider tracking. Only two of the six are shown to avoid redundancy, remaining four show very similar output.}
\label{fig:learned_pairwise_spider}
\end{figure}

The pairwise potential functions learned by DNBP in this task are visualized as was done in the double pendulum task. \cref{fig:learned_pairwise_spider} shows qualitative output from two of the six models. Only two are shown to avoid redundancy; chosen results are representative of remaining four potential functions. The left column of \cref{fig:learned_pairwise_spider} shows the normalized histogram of pairwise translations as computed from the training set for $X_1-X_0$ (top) and $X_4-X_1$ (bottom). The middle column of \cref{fig:learned_pairwise_spider} shows the normalized histogram of samples from learned pairwise sampler networks, $\psi_{0,1}^{\sim}(\cdot)$ and $\psi_{1,4}^{\sim}(\cdot)$. Finally in the right column of \cref{fig:learned_pairwise_spider}, uniformly sampled output ($100$x$100$ samples across pixel space) of the learned pairwise density networks is shown. Once again, the visual similarity between each learned potential function and the corresponding true distribution of pairwise translations is an indicator that DNBP is successful in learning to model each pairwise factor. Observe that the learned potential functions for $\psi_{1,4}(\cdot)$, which correspond to a revolute articulation, show no bias in favor of the downward configuration. This result is notably different from the potential functions learned on the double pendulum task and can be explained by the absence of gravity in the spider simulation. Similarly, the learned models for $\psi_{0,1}(\cdot)$ on the spider task exhibit a torus shape due to the effect of prismatic motion associated with the corresponding joint's articulation type and constraint. 

\section{Discussion}
\label{sec:discussion}

In this work, we proposed a novel formulation of belief propagation which is differentiable and uses a nonparametric representation of belief. It was hypothesized that combining maximum likelihood estimation with the nonparametric inference approach would enable end-to-end learning of the probabilistic factors needed for inference. The hypothesis was tested on both qualitative and quantitative experiments. Preliminary results demonstrate successful application of this approach. These results motivate further experiments, in particular a direct comparison with the differentiable particle filter to investigate the potential benefit of using a factored inference approach as well as a comparison with other learning approaches proposed for belief propagation.

A limitation of the current approach is the use of non-differentiable resampling~\cite{dpf:JonschkowskiRB18,diffresamp:abs-2004-11938}. This stops gradients from backpropagating beyond a single message-belief update. In the context of particle filtering, the use of non-differentiable resampling was addressed by Zhu et al. \cite{diffresamp:abs-2004-11938} using a learned \textit{particle transformer} network and also by Karkus et al. \cite{pfnet:KarkusHL18} using a differentiable \textit{soft-resampling} function. Exploration of these resampling approximations for DNBP is left as future work.

While this approach has reduced the demand for prior knowledge in the form of potential functions, DNBP still requires a graph as input. A direction for future work would attempt to learn the graphical model's factorization along with each potential function. We can envision a variant of DNBP which is initialized with a fully connected graph and learns to `cull' certain edges. This would further reduce the domain knowledge needed to apply nonparametric belief propagation.

\bibliographystyle{IEEEtran}
\bibliography{lbpbib}
\clearpage
\section{Appendix}
\subsection{Algorithm Pseudo Code}
\label{appendix:alg_pseudo}
In this section, pseudo code of the DNBP message passing algorithm is given for reference. As discussed in \cref{sec:dnbp}, this algorithm is a differentiable variant of the PMPNBP algorithm~\cite{bpposeest:DesinghLOJ19}. 

\newlength\myindent
\setlength\myindent{2em}
\newcommand\bindent{%
  \begingroup
  \setlength{\itemindent}{\myindent}
  \addtolength{\algorithmicindent}{\myindent}
}
\newcommand\eindent{\endgroup}

\begin{algorithm}[ht]
\SetAlgoLined
\SetKwInOut{Input}{input}\SetKwInOut{Output}{output}
\Input{Belief set $\bel{d}{n-1}=\set{(\wgt{d}{}{i},\prt{d}{}{i})}_{i=1}^{T}$\newline Incoming messages $\msg{u}{s}{n-1}=\set{(\wgt{u}{s}{i},\prt{u}{s}{i})}_{i=1}^{M}$ for each node $u\in \rho(s)\setminus d$}

\Output{Outgoing messages, $\msg{s}{d}{n}=\set{(\prt{s}{d}{i},\wgt{s}{d}{i})}_{i=1}^{M}$}
\BlankLine
 Draw $(1-\gamma^{n-1})\cdot M$ independent samples from $\bel{d}{n-1}$\newline$\set{\prt{s}{d}{i}\gets \bel{d}{n-1}}_{i=1}^{(1-\gamma^{n-1})\cdot M}$\;
 Apply particle diffusion to each sampled particle\newline$\prt{s}{d}{i}=\prt{s}{d}{i}+\temporal{d}$\;
 Draw remaining $\gamma^{n-1}\cdot M$ samples independently from uniform proposal distribution\;
 
 \ForEach{$\set{\prt{s}{d}{i}}_{i=1}^{M}$}{
    \For{$\ell=[1:U]$}{
    Sample $\hat{X_s}^{(i)}\sim \pairwise{s}{d}{\prt{s}{d}{i}}{}$\;
    $w_{unary}^{(i)} = w_{unary}^{(i)} + \unary{s}{\hat{X_s}^{(i)}}{}$\;
    }
    
    $w_{unary}^{(i)} = \frac{w_{unary}^{(i)}}{U}$\;
    
    \ForEach{$u\in \rho(s)\setminus d$}{
        $W_u^{(i)}=\sum\limits_{j=1}^{M}\wgt{u}{s}{j}\times w_{u}^{(ij)}$ where $w_{u}^{(ij)}=\pairwise{s}{d}{\prt{u}{s}{j}}{\prt{s}{d}{i}}$\;
    }
    $w_{neigh}^{(i)} = \prod\limits_{u\in \rho(s)\setminus d} W_u^{(i)}$\;
    $\wgt{s}{d}{i}=w_{unary}^{(i)}\times w_{neigh}^{(i)}$\;
 }
 
 Associate $\set{\wgt{s}{d}{i}}_{i=1}^{M}$ with $\set{\prt{s}{d}{i}}_{i=1}^{M}$ to form outgoing $\msg{s}{d}{n}$\;
 \caption{Message update}
\end{algorithm}

\begin{algorithm}[ht]
\SetAlgoLined
\SetKwInOut{Input}{input}\SetKwInOut{Output}{output}
\Input{Incoming messages, $\msg{s}{d}{n}=\set{(\wgt{s}{d}{i},\prt{s}{d}{i})}_{i=1}^{M}$, for each node $s\in \rho(d)$}
\Output{Belief set $\bel{d}{n}=\set{(\wgt{d}{}{i},\prt{d}{}{i})}_{i=1}^{T}$}
\BlankLine
\ForEach{$s\in \rho(d)$}{
 Update message weights \newline$\wgt{s}{d}{i} = \wgt{s}{d}{i} \times \unary{d}{\prt{s}{d}{i}}{}$ for $i\in[1:M]$\;

 Normalize message weights \newline$\wgt{s}{d}{i} = \frac{\wgt{s}{d}{i}}{\sum\limits_{j=1}^{M} \wgt{s}{d}{j}}$ for $i\in[1:M]$\;
 }
 Form belief set $\bel{d}{n}=\bigcup\limits_{s\in \rho(d)} \msg{s}{d}{n}$\;
 Normalize belief weights \newline$\wgt{d}{}{i} = \frac{\wgt{d}{}{i}}{\sum_{j=1}^{T} \wgt{d}{}{j}}$ for $i\in[1:T]$\;
 
 \caption{Belief update}
\end{algorithm}

Results in this work were generated with $U$ set to $10$, while past work~\cite{bpposeest:DesinghLOJ19} used $U=1$. This modification improved training stability during initial development. Note that $\gamma$ is a hyperparameter that controls the resampling strategy and is set to $0.9$ in our experiments. $\gamma$ is used only during training; during evaluation, all $M$ samples are drawn from $\bel{d}{n-1}$.

\subsection{Double Pendulum Clutter}
\label{appendix:pendulum_clutter}
As described in \cref{sec:experiment_pendulum}, the double pendulum dataset was generated using a modified version of the OpenAI~\cite{openai:1606.01540} Acrobot environment. Synthetic geometric shapes are rendered into each image of the dataset to simulate noisy, cluttered environments. All simulated clutter on the double pendulum task is generated according to the following parameters: $50\%$ of clutter is rendered visually beneath the pendulum while the remaining $50\%$ is rendered on top of the pendulum. For dynamic clutter, each geometry simulates motion using a random, constant position velocity ($\dot{x}$, $\dot{y}$) and orientation velocity ($\dot{\theta}$). Position velocities are sampled from $\mathcal{N}(0, 0.025)$. Orientation velocities are sampled from $\mathcal{N}(0, 0.05)$. Clutter is simulated as either rectangles with $80\%$ probability or circles with $20\%$ probability. Clutter rectangles are sized randomly with length of $\text{max}(0, l\sim\mathcal{N}(0.2, 0.05))$ and height of $\text{max}(0, h\sim\mathcal{N}(0.8, 0.2))$. Color of clutter rectangles is randomly chosen with RGB of $(0,204,204)$ or $(245,87,77)$. Clutter circles are sized randomly with radius of $\text{max}(0, r\sim\mathcal{N}(0.1, 0.1))$ and colored randomly with RGB of $(204,204,0)$ or $(96,217,63)$. Size and color distributions were chosen to ensure clutter visually resembles the double pendulum parts. The position of each clutter geometry was randomly initialized within $1.5$x the extent of the image boundary.

The training and validation datasets were distributed evenly among clutter ratios of [$0$, $0-0.04$, and $0.04-0.1$]. For the training/validation sequences that included clutter, the number of clutters rendered beneath and on top of the double pendulum was individually randomly sampled from independent Binomial distributions using $n=15$, $p=0.3$. To generate the test set, which was uniformly distributed among clutter ratios as described in \cref{sec:experiment_pendulum}, rejection sampling was used with variable numbers of rendered geometries.

\subsection{Articulated Spider Model}
\label{appendix:spider_model}
Data for the articulated spider tracking task of \cref{sec:experiment_spider} was simulated using the Pillow~\cite{clark2015pillow} image processing library. Three revolute-prismatic joints are all centrally located and treated as the root of the spider's kinematic tree. The remaining three revolute joints are attached to pairs of links, forming three distinct 'arms' of the spider. Each joint is rendered as a yellow circle while the six rigid-body links are rendered as distinct red, green or blue rectangles respectively. Size parameters that follow are with respect to rendered image size of  $500$x$500$px. The three inner revolute-prismatic joints include rotational constraints limiting each to a non-overlapping $120^{\circ}$ range of articulation as well as prismatic constraints limiting the extension to within $[20,80]$ pixels of translation. The three purely revolute joints are constrained to rotations between $\pm35^{\circ}$ with respect to their local origins. Each rigid-body link has width of $20$px and height of $80$px pixels while each joint has radius of $10$px. 

For every simulated sequence, the spider is initialized with uniformly random root position within the central $180$x$180$px window and uniformly random root orientation from [$0,2\pi$]. Furthermore, each joint state is initialized uniformly at random within its particular articulation constraints. The spider is simulated with dynamics using randomized, constant root, and joint velocities with respect to a time step ($dt$) of $0.01$. The root's position velocities ($\dot{x}$, $\dot{y}$) are each sampled from an equally weighted $2$-component Gaussian mixture with means ($+24$, $-24$) and standard deviations ($15$, $15$). Whereas, the root's orientation velocity ($\dot{\theta}$) is sampled from an equally weighted $2$-component Gaussian mixture with means ($+0.3$, $-0.3$) and standard deviations ($0.1$, $0.1$). Each rotational joint's velocity is sampled from an equally weighted $2$-component Gaussian mixture with means ($+0.3$, $-0.3$) and standard deviations ($0.1$, $0.1$). Similarly, each prismatic joint's velocity is sampled from an equally weighted $2$-component Gaussian mixture with means ($+500$, $-500$) and standard deviations ($60$, $60$). Note that if any joint reaches an articulation limit during simulation, the direction of its velocity is reversed.

\subsection{Articulated Spider Clutter}
\label{appendix:spider_clutter}
Clutter generation for the articulated spider tracking task follows a similar generation process as was used for the double pendulum task. Clutter parameters that follow are with respect to rendered image size of $500$x$500$px and time step ($dt$) of $0.01$. $50\%$ of clutter is rendered beneath and $50\%$ is rendered on top of the spider. For dynamic clutter, each geometry simulates motion using a random, constant position velocity ($\dot{x}$, $\dot{y}$) and orientation velocity ($\dot{\theta}$). Position velocities are sampled from $\mathcal{N}(0, 3)$ while orientation velocities are sampled from $\mathcal{N}(0, 0.05)$. Clutter is simulated as either a rectangle with $70\%$ probability or a circle with $30\%$ probability. Clutter rectangles are sized randomly with length of $\text{max}(0, l\sim\mathcal{N}(20, 3))$ and height of $\text{max}(0, h\sim\mathcal{N}(80, 5))$. The color of clutter rectangles is chosen uniformly at random from the same colors as were used for the spider arms. Clutter circles are sized randomly with a radius of $\text{max}(0, r\sim\mathcal{N}(10, 3))$ and colored yellow to match the color of the spider's joints. The position of each clutter geometry was randomly initialized within the image boundary.

For the training/validation sequences that included clutter, the number of clutter shapes rendered beneath and on top of the double pendulum was each randomly sampled from independent Binomial distributions using $n=10$, $p=0.5$. The test set was generated with uniformly distributed clutter ratios, as described in \cref{sec:experiment_spider}, using rejection sampling with variable numbers of rendered geometries.

\end{document}